# RESEARCH AND APPLICATION OF TIME SERIES ALGORITHMS IN CENTRALIZED PURCHASING DATA

**Yun Bai, Suling Jia, Xixi Li**
**School of Economics and Management, Beihang University, Beijing 100191, China**
**E-mail: 1508920895@qq.com**

## Abstract

Based on the online transaction data of COSCO group's centralized procurement platform, this paper studies the clustering method of time series type data. The different methods of similarity calculation, different clustering methods with different K values are analysed, and the best clustering method suitable for centralized purchasing data is determined. The company list under the corresponding cluster is obtained. The time series motif discovery algorithm is used to model the centroid of each cluster. Through ARIMA method, we also made 12 periods of prediction for the centroid of each category. This paper constructs a matrix of "Customer Lifecycle Theory - Five Elements of Marketing ", and puts forward corresponding marketing suggestions for customers at different life cycle stages.

**Keywords:** Time Series, Clustering, Purchasing Behavior, Customer Lifecycle

## Introduction

COSCO is one of the largest shipping companies in mainland China, one of the largest state-owned enterprises directly under the central government of China, and one of the largest marine transport company in the world. ZNYX is COSCO's online centralized procurement platform for office supplies, which is linked to the ChenGuang, SuNing and other suppliers and subsidiary of COSCO. The buyer companies order on the platform, which is responsible for data management. Suppliers select and distribute goods, and the group achieves targets of unified control.

COSCO is very concerned about the customer value of the buyer companies. We can understand the consumption of office supplies by analyzing the purchasing behavior patterns of buyer companies. At the same time, it also reflects the business situation of the company. The ZNYX platform began to be used in April 2017. As of April 2018, more than ten thousand orders are generated, including 399 participating companies. For each buyer company, their purchase amounts are arranged to form time series according to time order. Through further sorting, the transaction record data of the platform can be collated into a time series data set. The data of all companies are unmarked. To judge the type of buying behavior of a company, we first turn the problem into a time series clustering analysis.

## Similarity measurement

The core step of clustering analysis is similarity measurement. In the field of similarity measurement, the distance between two time series is usually used to measure their similarity. The common distance calculation methods are: Euclidean distance(EUCL); the DTW distance (DTW) used by Berndt and Clifford in the field of financial pattern recognition[1]; the distance based on the correlation coefficient (COR) proposed by Golay[2]; the distance based on the autocorrelation coefficient (ACF) defined by Galeano and Pena[3]; the distance based on the periodic graph (PER) introduced by Caiado[4]; Casado de Lucas takes into account the cumulative form of periodogram (INT.PER), such as integral periodogram; Fan and Kreutzberger have been revised based on the periodogram, proposed non parametric spectral estimation based on distance (SPEC.GLK) [5]. This paper will use these distance calculation methods to calculate with different clustering algorithms.

## Clustering Algorithm

There are three kinds of clustering algorithms in this paper: hierarchical clustering, partitional clustering and fuzzy clustering. In the clustering algorithm, through similarity measurement, similar data are divided into one class, and dissimilar data are divided into different categories. Finally, a number of clusters are obtained, in which the data are similar, and among which the data are dissimilar.

## Clustering Validity Index

There are 7 similarity measurement algorithms and 3 clustering algorithms are introduced. If these algorithms are arranged and combined, 21 clustering schemes will be generated. Obviously, this is too much for this research. Therefore, clustering validity index need to be introduced to evaluate various schemes for COSCO centralized purchase data set.

Generally speaking, the clustering validity index only evaluates the results, ignoring the working principle and classification of the algorithm itself. The clustering validity indexes can be divided into two categories according to the different methods: internal CVI, such as Silhouette index[6]; and external CVI[7], such as information variation index. The common defects of the two indexes are that they can not be used in fuzzy clustering. In 2001, Kalpakis et al put forward Sim index to measure the similarity of two clustering results, which can be applied to hierarchical clustering, partitional clustering and fuzzy clustering[8].

**Dimensionality Reduction**

Time series dimensionality reduction algorithm can reduce the high-dimensional data to two-dimensional plane. In this paper, the t-SNE reduction algorithm is used, and the t-SNE algorithm is a nonlinear, random nearest neighbor embedding algorithm based on t distribution, and the structure of the internal data is found by random walk between adjacent data.

**Motif Discovery Algorithm**

Time series motif discovery is an important part of timing analysis. Motif discovery algorithm aims to find a similar or identical pattern segment in a sequence. In 2003, Chiu and others proposed a single variable motif mining method [9]. In 2009, Vahdatpour and others proposed a multivariable motif mining method.[10]

**ARIMA Forecasting Model**

ARIMA (Autoregressive Integrated Moving Average model) is one of the time series forecasting methods. In ARIMA (p, d, q), AR is "autoregressive" and P is the number of autoregressive terms; MA is "sliding average" and Q is the number of sliding average terms; D is the difference number (order) made to make it a stationary sequence.

**Customer Lifecycle value theory (CLV) and Five Elements of Marketing (FEM)**

For an enterprise, its customers often go through five stages: acquisition, promotion, maturity, recession and departure. In the different life stages of the customer, the focus of enterprise is different. In the acquisition stage, the enterprise pays attention to how to find and obtain the potential customers; in the stage of promotion, the enterprise pays attention to how to train customers into high value customers; in the mature stage, the enterprise pays attention to how to foster customer loyalty; in the recession stage, enterprise is concerned about how to extend the customer life cycle; in the stage of departure, enterprises are concerned about how to win back customers.

In the field of marketing, there are five factors worth considering: market segmentation, product combination/packaging, channel combination, targeted advertising/promotion, loyalty management.

## Analysis and Solution

**Experimental scheme**

1. Preprocess the experimental data, obtain the total time series of the total purchases of 399 buyer companies, and take weeks as time units.
2. Set K different cluster numbers, and calculate the Sim value of hierarchical clustering, partitional clustering and fuzzy clustering when the distance calculation method is different.
3. Select the combination scheme with the maximum Sim value and get the list of buyer companies under each cluster.
4. T-SNE algorithm is used to reduce the dimensionality of time series data of the above scheme.
5. The mean time series of each cluster is calculated as centroid, and the centroid is used for pattern discovery algorithm.
6. The results of pattern discovery are classified according to the customer life cycle theory.
7. Construct "CLV-FEM" matrix, and put forward marketing suggestions for users at all stages of life.

**Data Preprocessing**

The dataset studied in this paper comes from all transaction records on the ZNYX platform, from May 2017 to April 2018. First, we get the centralized procurement transaction data from the database. The transaction record contains 13391 lines and 80 columns of data. We only choose three columns related to time series: REAL_PRICE, CREATE_TIME and ORGFULLNAME. In addition, the purchase of commodity names, purchase quantities, tax rates, suppliers, etc. are not considered here.

Then the data of CREATE_TIME is processed, which mainly includes sorting the time and changing the format from the time stamp to the string form.

After that, the data were aggregated by company, then we get the transaction of 399 buyer companies. We collate the time in weeks, and get a general table of all company time series. Each row in the table represents a company's time series of total amount of purchase.

Finally, all the data are converted to the numeric format. The part of the preprocessing ends at this point. The time series of each company is shown in Figure 1.

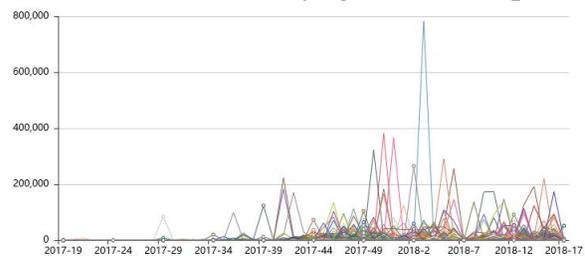

Figure 1 Time series of total amount of purchase of each company

**Similarity measurement**

Based on the 7 distance calculation methods mentioned above, the distance between the time series of each company is calculated in this paper. The calculation result is a 399 * 399 matrix. Due to the large number of companies, we only choose the distance between the top 6 companies to display. The results are attached to the appendices.

**Cluster Analysis**

In this part, three clustering algorithms are used to calculate and analyze the data sets. The initial cluster number K ranges from 2 to 10, and different Sim values are obtained by using different distance calculation methods. The maximum K value corresponds to the best plan we expect.

Table 1 Sim value table of hierarchical clustering

| Num of K | EUCL | DTW | COR | ACF |
|---|---|---|---|---|
| K=2 | 0.007525062 | 0.007525062 | 0.009990986 | 0.009975816 |
| K=3 | 0.01003759 | 0.01154503 | 0.01494331 | 0.01477023 |
| K=4 | 0.01380634 | 0.015565 | 0.01988938 | 0.01937601 |
| K=5 | 0.01631887 | 0.01891502 | 0.02475144 | 0.02368293 |
| K=6 | 0.0188314 | 0.02159509 | 0.02956826 | 0.0279937 |
| K=7 | 0.02134393 | 0.0238564 | 0.03439234 | 0.03175855 |
| K=8 | 0.0253639 | 0.02762514 | 0.03928855 | 0.03594862 |
| K=9 | 0.02871392 | 0.02971894 | 0.04414581 | 0.04071231 |
| K=10 | 0.03097522 | 0.03411568 | **0.04885244** | 0.0431868 |
| Num of K | PACF | PER | INT.PER | SPEC.GLK |
| K=2 | 0.009947441 | 0.007525062 | 0.009939174 | 0.009999676 |
| K=3 | 0.01488835 | 0.01003759 | 0.0137021 | 0.01376819 |
| K=4 | 0.01955482 | 0.01338761 | 0.01865373 | 0.01870943 |
| K=5 | 0.02448277 | 0.01740758 | 0.02356376 | 0.02327615 |
| K=6 | 0.02912978 | 0.01908263 | 0.02757475 | 0.02802235 |
| K=7 | 0.03383297 | 0.02310259 | 0.03187565 | 0.03294035 |
| K=8 | 0.03851254 | 0.0253639 | 0.03637035 | 0.03503415 |
| K=9 | 0.04269966 | 0.02804397 | 0.040544 | 0.03995244 |
| K=10 | 0.04728632 | 0.0322314 | 0.04534465 | 0.04467127 |

Table 2 Sim value table of partitional clustering

| Num of K | SBD | DTW | SDTW |
|---|---|---|---|
| K=2 | 0.009945909 | 0.009951719 | 0.009998843 |
| K=3 | 0.01474662 | 0.01483422 | 0.0149549 |
| K=4 | 0.01959149 | 0.01963765 | 0.01988796 |
| K=5 | 0.02451691 | 0.02357098 | 0.02480725 |
| K=6 | 0.02914283 | 0.02904528 | 0.02968041 |
| K=7 | 0.03392118 | 0.03359927 | 0.03440889 |
| K=8 | 0.03791502 | 0.03703196 | 0.03920895 |
| K=9 | 0.04271304 | 0.04200897 | 0.04405334 |

| K=10 | 0.04727171 | **0.04467682** | 0.0487271 |

Table 3 Sim value table of fuzzy clustering

| Num of K | EUCL | DTW | SBD |
|---|---|---|---|
| K=2 | 0.00999521 | 0.009995386 | 0.009990986 |
| K=3 | 0.01492013 | 0.01492549 | 0.01491542 |
| K=4 | 0.01931819 | 0.01975013 | 0.01985357 |
| K=5 | 0.02266688 | 0.02423782 | 0.0245095 |
| K=6 | 0.02704946 | 0.02872649 | 0.02860986 |
| K=7 | 0.03109042 | 0.03115708 | 0.03328198 |
| K=8 | 0.03533632 | 0.03795387 | 0.03731916 |
| K=9 | 0.03033411 | 0.03783905 | 0.03794477 |
| K=10 | 0.03536026 | **0.0426304** | 0.04238671 |

By comparison, we find that when K=10 and COR distance is used, the Sim value is the largest, 0.04885244. It is proved that the combination of these algorithms can produce the best clustering results for the dataset in this paper.

The results of hierarchical clustering are shown as follows：

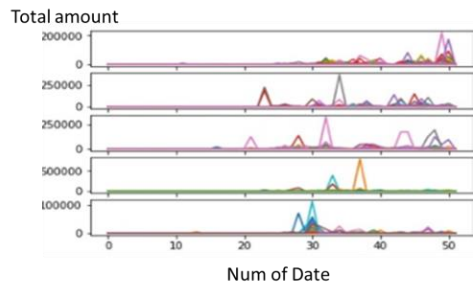

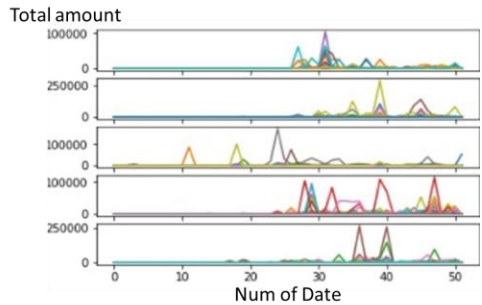

Figure 3 Time series set of 10 clusters

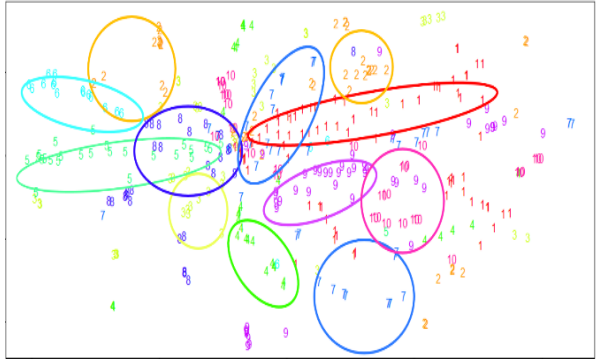

Figure 4 The result of dimensionality reduction

**Motif Discovery**

After clustering analysis, we can not distinguish the clustering results by intuition alone. Obviously, the data in each cluster are very similar, and the centroid is a concentrated expression of this similarity. Therefore, we find out the implicit rules of the centroid application motif discovery algorithm, and this law also reflects the similar purchase behavior of the companies in this category.

Then we classify the results of motif discovery according to the CLV theory. The result is as follows:

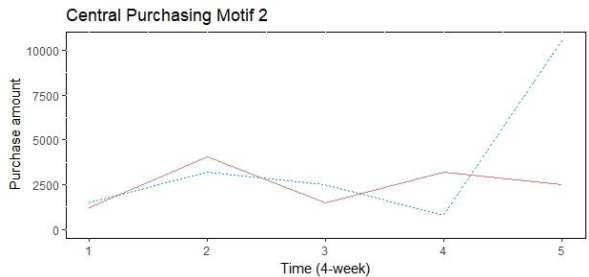

Figure 5 Motif of Type 1

The first type of users began to purchase late. There is a certain fluctuation in the early stage, and then it increases significantly later. It is considered that this type of user belongs to the stage of **acquisition**.

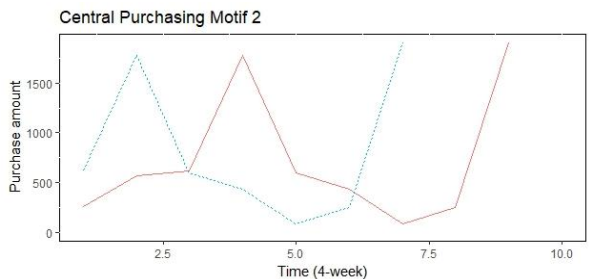

Figure 6 Motif of Type 8

The purchasing behavior of the eighth types of users has a distinct upward trend, and it can be considered that they are in the **promotion** stage of the customer lifecycle.

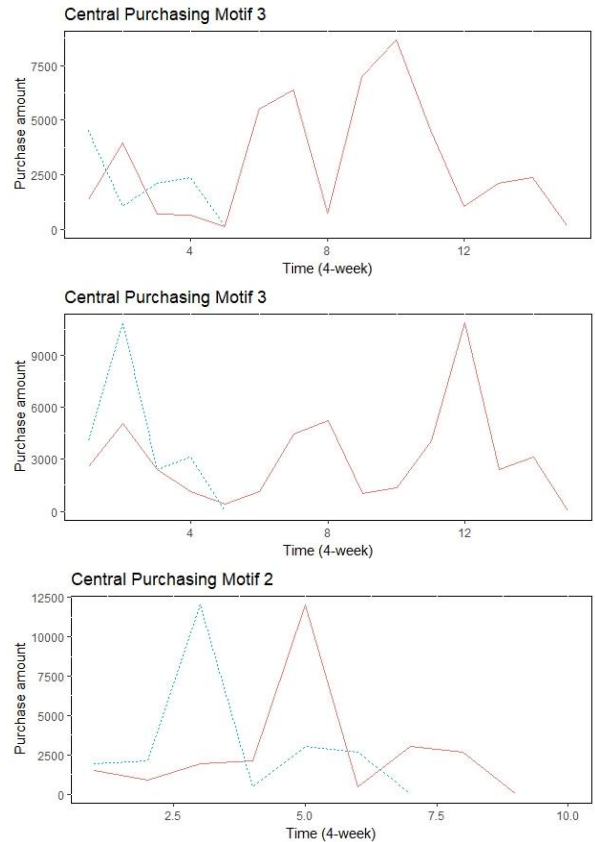

Figure 7 Motif of Type 2, 3, 9

These three types of users began to purchase earlier, and have larger purchase amount and more frequent procurement activities, which could be considered as a stage of **maturity**.

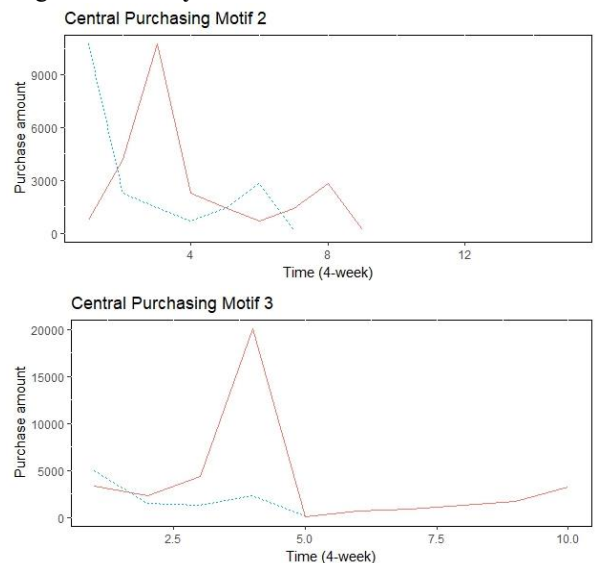

Figure 8 Motif of Type 7, 10

These companies have had excessive amount of buying behaviors, but the amount and frequency of purchases have tended to decrease. It can be considered that such companies are in the **recession** stage.

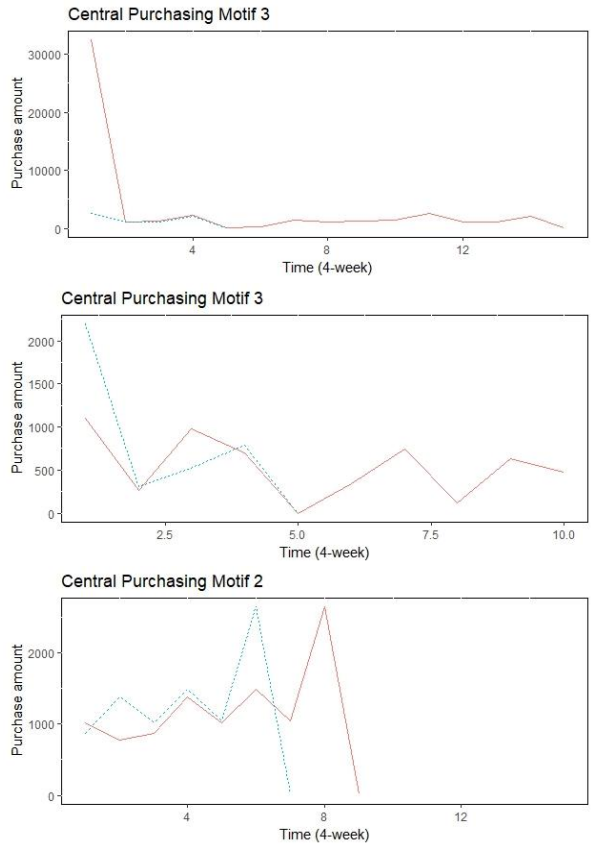

Figure 9 Motif of Type 4, 5, 6

At the end of 2017, these companies had a large amount of purchase behavior, and after that they almost stopped buying or buying too little, which could be considered to be a departure stage.

Then, we use ARIMA algorithm to predict the centroid of each class in 3 periods. The model and AIC are shown in the table below:

Table 4 ARIMA models and AIC scores of different clusters

| Categories | Model | AIC | Prediction 1 | Prediction 2 | Prediction 3 |
|---|---|---|---|---|---|
| clust1 | ARIMA(2,0,2) | 906.45 | 4210.321 | 9447.201 | 3696.422 |
| clust2 | ARIMA(3,0,0) | 951.93 | 1417.971 | 1659.643 | 1180.471 |
| clust3 | ARIMA(3,0,1) | 956.54 | 1796.8468 | 888.1346 | 2080.97 |
| clust4 | ARIMA(3,0,0) | 1039.7 | 1870.321 | 1896.71 | 1872.653 |
| clust5 | ARIMA(0,0,1) | 977.31 | 794.6652 | 901.5987 | 901.5987 |
| clust6 | ARIMA(0,0,1) | 987.17 | 1051.653 | 1225.92 | 1225.92 |
| clust7 | ARIMA(2,0,2) | 963.64 | 1620.877 | 1887.911 | 1564.774 |
| clust8 | ARIMA(0,0,1) | 877.23 | 1005.7126 | 780.7293 | 780.7293 |
| clust9 | ARIMA(0,0,1) | 943.36 | 1008.11 | 1190.505 | 1190.505 |
| clust10 | ARIMA(3,0,0) | 983.85 | 1235.273 | 1530.8 | 1231.252 |

Next, we construct a "CLV-FEM" matrix, and put forward marketing suggestions according to different marketing elements for companies in different life stages.

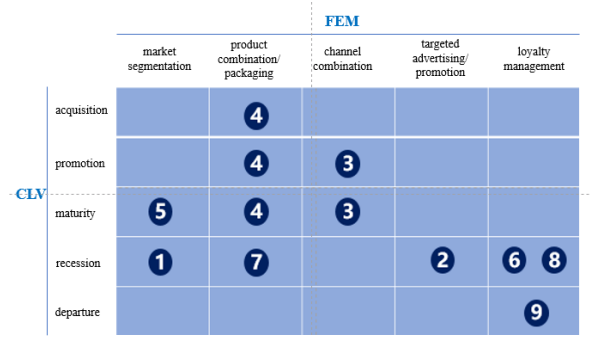

Figure 10 "CLV-FEM" matrix

The suggestions for the corresponding numbers in the graph are as follows:

1. Establish regular customer churn system for high value customers.
2. Study the causes of customer churn in detail and carry out corresponding promotional work.
3. Combine different marketing channels to enhance targeted marketing capabilities, such as online marketing and offline marketing.
4. According to the relevance of purchasing office supplies, the related products are bundled together to form a product package: A4 paper + printer, stapler + staples, etc.
5. Classify purchasing groups according to their commercial value. For high valued users, we can improve their loyalty by adopting regular care and tracking purchase changes.
6. Analyze the reasons why users are no longer interested in the platform, and establish early warning systems.
7. When a high-value customer is in recession, we can design product packages to retain them, such as printer + toner cartridge.
8. Set up the database of the customers leaving the stage, including their contact information, etc., so as to facilitate the detailed investigation of the reasons for their leaving.
9. Regularly visit the high value users who have left, analyze their reasons for leaving, understand the probability of recovering them, and provide reference materials for the establishment of early warning system.

## Conclusions

In this paper, a large number of algorithms have been studied. Through Sim index calculation, we have found the most suitable clustering solution for centralized purchasing data of office supplies. After clustering and discovering the time series of the companies in the dataset, all of the companies are classified into different life stages according to the customer lifecycle value theory. The innovation of this paper is that we have constructed the "CLV-FEM" matrix, and put forward marketing suggestions to suppliers in view of the different life stages of the company. The study of this paper has certain commercial value.